\documentclass[conference]{IEEEtran}
\IEEEoverridecommandlockouts
\usepackage{amsmath,amssymb,amsfonts}
\usepackage{algorithm}
\usepackage[noend]{algpseudocode}
\usepackage{amsmath}
\usepackage{graphicx}
\usepackage{subcaption}
\usepackage{textcomp}
\usepackage{xcolor}
\usepackage{acronym}
\usepackage{siunitx}
\usepackage{hyperref}
\definecolor{belize}{RGB}{41,128,185}
\definecolor{pomgrenate}{RGB}{192, 57, 43}

\hypersetup{colorlinks=True,linkcolor={blue},citecolor={blue},urlcolor={blue}}

\usepackage[backend=biber, style=ieee,sorting=none,maxnames=2,
minnames=1, pluralothers=true,isbn=false,eprint=false,doi=true,url=false,
arxiv=abs,
]{biblatex}

\definecolor{comment}{RGB}{1, 163, 164}

\def\BibTeX{{\rm B\kern-.05em{\sc i\kern-.025em b}\kern-.08em
    T\kern-.1667em\lower.7ex\hbox{E}\kern-.125emX}}

\acrodef{ADC}[ADC]{Analog-to-Digital Converter}
\acrodef{ADEXP}[AdExp-IF]{Adaptive Exponential Integrate-and-Fire}
\acrodef{ADM}[ADM]{Asynchronous Delta Modulator}
\acrodef{AE}[AE]{Address-Event}
\acrodef{AER}[AER]{Address-Event Representation}
\acrodef{AEX}[AEX]{AER EXtension board}
\acrodef{AFE}[AFE]{Analog Front-End}
\acrodef{AFM}[AFM]{Atomic Force Microscope}
\acrodef{AGC}[AGC]{Automatic Gain Control}
\acrodef{AI}[AI]{Artificial Intelligence}
\acrodef{alphaMN}[\( \alpha \)-MN]{$\alpha$-motor neuron}
\acrodef{AMDA}[AMDA]{AER Motherboard with D/A converters}
\acrodef{AMPA}[AMPA]{$\alpha$-Amino-3-hydroxy-5-methyl-4-isoxazolepropionic Acid}
\acrodef{ANN}[ANN]{Artificial Neural Network}
\acrodef{API}[API]{Application Programming Interface}
\acrodef{APMOM}[APMOM]{Alternate Polarity Metal On Metal}
\acrodef{ARM}[ARM]{Advanced RISC Machine}
\acrodef{ASIC}[ASIC]{Application Specific Integrated Circuit}
\acrodef{BCM}[BMC]{Bienenstock-Cooper-Munro}
\acrodef{BD}[BD]{Bundled Data}
\acrodef{BEOL}[BEOL]{Back-end of Line}
\acrodef{BG}[BG]{Bias Generator}
\acrodef{BMI}[BMI]{Brain-Machince Interface}
\acrodef{BTB}[BTB]{Band-to-Band tunnelling}
\acrodef{bpm}[bpm]{Beats per Minute}
\acrodef{CA}[CA]{Cortical Automaton}
\acrodef{CAD}[CAD]{Computer Aided Design}
\acrodef{CAM}[CAM]{Content Addressable Memory}
\acrodef{CAVIAR}[CAVIAR]{Convolution AER Vision Architecture for Real-Time}
\acrodef{CCN}[CCN]{Cooperative and Competitive Network}
\acrodef{CDR}[CDR]{Clock-Data Recovery}
\acrodef{CFC}[CFC]{Current to Frequency Converter}
\acrodef{CHP}[CHP]{Communicating Hardware Processes}
\acrodef{CMIM}[CMIM]{Metal-Insulator-Metal Capacitor}
\acrodef{CML}[CML]{Current Mode Logic}
\acrodef{CMOL}[CMOL]{Hybrid CMOS nanoelectronic circuits}
\acrodef{CMOS}[CMOS]{Complementary Metal-Oxide-Semiconductor}
\acrodef{CNN}[CNN]{Convolutional Neural Network}
\acrodef{CNS}[CNS]{central Nervous System}
\acrodef{COTS}[COTS]{Commercial Off-The-Shelf}
\acrodef{CPG}[CPG]{Central Pattern Generator}
\acrodef{CPLD}[CPLD]{Complex Programmable Logic Device}
\acrodef{CPU}[CPU]{Central Processing Unit}
\acrodef{CSM}[CSM]{Cortical State Machine}
\acrodef{CSP}[CSP]{Constraint Satisfaction Problem}
\acrodef{CTXCTL}[CTXCTL]{CortexControl}
\acrodef{CV}[CV]{Coefficient of Variation}
\acrodef{DAC}[DAC]{Digital to Analog Converter}
\acrodef{DAS}[DAS]{Dynamic Auditory Sensor}
\acrodef{DAVIS}[DAVIS]{Dynamic and Active Pixel Vision Sensor}
\acrodef{DBN}[DBN]{Deep Belief Network}
\acrodef{DBS}[DBS]{Deep Brain Stimulation}
\acrodef{DFA}[DFA]{Deterministic Finite Automaton}
\acrodef{DIBL}[DIBL]{Drain-Induced Barrier-Lowering}
\acrodef{DI}[DI]{Delay Insensitive}
\acrodef{divmod3}[DIVMOD3]{Divisibility of a number by three}
\acrodef{DL}[DL]{Deep Learning}
\acrodef{DMA}[DMA]{Direct Memory Access}
\acrodef{DNF}[DNF]{Dynamic Neural Field}
\acrodef{DNN}[DNN]{Deep Neural Network}
\acrodef{DoA}[DoA]{Degree of Actuation}
\acrodef{DoF}[DoF]{Degree of Freedom}
\acrodef{DPE}[DPE]{Dynamic Parameter Estimation}
\acrodef{DPI}[DPI]{Differential Pair Integrator}
\acrodef{DRAM}[DRAM]{Dynamic Random Access Memory}
\acrodef{DR}[DR]{Dual Rail}
\acrodef{DRRZ}[DR-RZ]{Dual-Rail Return-to-Zero}
\acrodef{DSP}[DSP]{Digital Signal Processor}
\acrodef{DVS}[DVS]{Dynamic Vision Sensor}
\acrodef{DYNAP-SE}[DYNAP-SE]{Dynamic Neuromorphic Asynchronous Processor}
\acrodef{EBL}[EBL]{Electron Beam Lithography}
\acrodef{ECG}[ECG]{Electrocardiography}
\acrodef{ECoG}[ECoG]{Electrocorticography}
\acrodef{EDA}[EDA]{Electrodermal activity}
\acrodef{EDVAC}[EDVAC]{Electronic Discrete Variable Automatic Computer}
\acrodef{EEG}[EEG]{Electroencephalography}
\acrodef{EI}[EI]{Excitatory-Inhibitory}
\acrodef{EIN}[EIN]{Excitatory-Inhibitory Network}
\acrodef{EM}[EM]{Expectation Maximization}
\acrodef{EMG}[EMG]{Electromyography}
\acrodef{EOG}[EOG]{Electrooculogram}
\acrodef{EPSC}[EPSC]{Excitatory Post-Synaptic Current}
\acrodef{EPSP}[EPSP]{Excitatory Post-Synaptic Potential}
\acrodef{EZ}[EZ]{Epileptogenic Zone}
\acrodef{FDSOI}[FDSOI]{Fully-Depleted Silicon on Insulator}
\acrodef{FET}[FET]{Field-Effect Transistor}
\acrodef{FFT}[FFT]{Fast Fourier Transform}
\acrodef{FI}[F-I]{Frequency--Current}
\acrodef{fastICA}[fastICA]{fast Independent Component Analysis}
\acrodef{FMA}[FMA]{Floating Microelectrode Array}
\acrodef{FNN}[FNN]{Feed-forward Neural Network}
\acrodef{FPGA}[FPGA]{Field Programmable Gate Array}
\acrodef{FR}[FR]{Fast Ripple}
\acrodef{FSA}[FSA]{Finite State Automaton}
\acrodef{FSM}[FSM]{Finite State Machine}
\acrodef{GABA}[GABA]{$\gamma$-Aminobutanoic Acid}
\acrodef{gammaMN}[\( \gamma \)-MN]{$\gamma$-motor neuron}
\acrodef{GIDL}[GIDL]{Gate-Induced Drain Leakage}
\acrodef{GOPS}[GOPS]{Giga-Operations per Second}
\acrodef{GPIO}[GPIO]{General Purpose I/O}
\acrodef{GPU}[GPU]{Graphical Processing Unit}
\acrodef{GT}[GT]{Ground Truth}
\acrodef{GUI}[GUI]{Graphical User Interface}
\acrodef{HAL}[HAL]{Hardware Abstraction Layer}
\acrodef{HFO}[HFO]{High Frequency Oscillation}
\acrodef{HH}[H\&H]{Hodgkin \& Huxley}
\acrodef{HMM}[HMM]{Hidden Markov Model}
\acrodef{HR}[HR]{Heart Rate}
\acrodef{HRS}[HRS]{High-Resistive State}
\acrodef{HSD}[HSD]{Honest Significant Difference} 
\acrodef{HSE}[HSE]{Handshaking Expansion}
\acrodef{HW}[HW]{Hardware}
\acrodef{hWTA}[hWTA]{Hard Winner-Take-All}
\acrodef{HRV}[HRV]{hearth rate variability}
\acrodef{Hyser}[Hyser]{High-densitY Surface Electromyogram}
\acrodef{IC}[IC]{Integrated Circuit}
\acrodef{ICA}[ICA]{Independent Component Analysis}
\acrodef{ICT}[ICT]{Information and Communication Technology}
\acrodef{iEEG}[iEEG]{Intracranial Electroencephalography}
\acrodef{IF2DWTA}[IF2DWTA]{Integrate \& Fire 2-Dimensional WTA}
\acrodef{IF}[I\&F]{Integrate-and-Fire}
\acrodef{IFSLWTA}[IFSLWTA]{Integrate \& Fire Stop Learning WTA}
\acrodef{IMU}[IMU]{Inertial Measurement Unit}
\acrodef{INCF}[INCF]{International Neuroinformatics Coordinating Facility}
\acrodef{INI}[INI]{Institute of Neuroinformatics}
\acrodef{IO}[I/O]{Input/Output}
\acrodef{IoT}[IoT]{Internet of Things}
\acrodef{IP}[IP]{Intellectual Property}
\acrodef{IPSC}[IPSC]{Inhibitory Post-Synaptic Current}
\acrodef{IPSP}[IPSP]{Inhibitory Post-Synaptic Potential}
\acrodef{ISI}[ISI]{Inter-Spike Interval}
\acrodef{JFLAP}[JFLAP]{Java - Formal Languages and Automata Package}
\acrodef{LEDR}[LEDR]{Level-Encoded Dual-Rail}
\acrodef{LFP}[LFP]{Local Field Potential}
\acrodef{LIFE}[LIFE]{Longitudinal Intrafascicular Electrodes}
\acrodef{LIF}[LIF]{Leaky Integrate-and-Fire}
\acrodef{LLC}[LLC]{Low Leakage Cell}
\acrodef{LMS}[LMS]{Least Mean Squares}
\acrodef{LNA}[LNA]{Low-Noise Amplifier}
\acrodef{LPF}[LPF]{Low Pass Filter}
\acrodef{LR}[LR]{Logistic Regression}
\acrodef{LRS}[LRS]{Low-Resistive State}
\acrodef{LSM}[LSM]{Liquid State Machine}
\acrodef{LTD}[LTD]{Long Term Depression}
\acrodef{LTI}[LTI]{Linear Time-Invariant}
\acrodef{LTP}[LTP]{Long Term Potentiation}
\acrodef{LTU}[LTU]{Linear Threshold Unit}
\acrodef{LUT}[LUT]{Look-Up Table}
\acrodef{LVDS}[LVDS]{Low Voltage Differential Signaling}
\acrodef{MCMC}[MCMC]{Markov-Chain Monte Carlo}
\acrodef{MAE}[MAE]{Mean Absolute Error}
\acrodef{MEA}[MEA]{Multielectrode Arrays}
\acrodef{MEMS}[MEMS]{Micro Electro Mechanical System}
\acrodef{MFR}[MFR]{Mean Firing Rate}
\acrodef{MIM}[MIM]{Metal Insulator Metal}
\acrodef{ML}[ML]{Machine Learning}
\acrodef{MLP}[MLP]{Multilayer Perceptron}
\acrodef{monoNSM}[monoNSM]{Monotonic Neural State Machine}
\acrodef{MOSCAP}[MOSCAP]{Metal Oxide Semiconductor Capacitor}
\acrodef{MOSFET}[MOSFET]{Metal Oxide Semiconductor Field-Effect Transistor}
\acrodef{MOS}[MOS]{Metal Oxide Semiconductor}
\acrodef{MRI}[MRI]{Magnetic Resonance Imaging}
\acrodef{MVC}[MVC]{Maximum Voluntary Contraction}
\acrodef{NCS}[NCS]{Neuromorphic Cognitive Systems}
\acrodef{NDFSM}[NDFSM]{Non-deterministic Finite State Machine}
\acrodef{ND}[ND]{Noise-Driven}
\acrodef{NEF}[NEF]{Neural Engineering Framework}
\acrodef{NHML}[NHML]{Neuromorphic Hardware Mark-up Language}
\acrodef{NIL}[NIL]{Nano-Imprint Lithography}
\acrodef{NI}[NI]{Neural Interface}
\acrodef{NMDA}[NMDA]{\textit{N}-Methyl-\textsc{d}-aspartate}
\acrodef{NME}[NE]{Neuromorphic Engineering}
\acrodef{NN}[NN]{Neural Network}
\acrodef{nnNSM}[nnNSM]{Nearest Neighbors Neural State Machine}
\acrodef{NOC}[NoC]{Network-on-Chip}
\acrodef{NRZ}[NRZ]{Non-Return-to-Zero}
\acrodef{NSM}[NSM]{Neural State Machine}
\acrodef{OR}[OR]{Operating Room}
\acrodef{OTA}[OTA]{Operational Transconductance Amplifier}
\acrodef{PCB}[PCB]{Printed Circuit Board}
\acrodef{PCHB}[PCHB]{Pre-Charge Half-Buffer}
\acrodef{PCM}[PCM]{Phase Change Memory}
\acrodef{PC}[PC]{Personal Computer}
\acrodef{PDK}[PDK]{Process Design Kit}
\acrodef{PE}[PE]{Phase Encoding}
\acrodef{PFA}[PFA]{Probabilistic Finite Automaton}
\acrodef{PFC}[PFC]{Prefrontal Cortex}
\acrodef{PFM}[PFM]{Pulse Frequency Modulation}
\acrodef{PGA}[PGA]{Programmable Gain Amplifier}
\acrodef{PID}[PID]{Proportional–integral–derivative}
\acrodef{PNI}[PNI]{Peripheral Nerve Interface}
\acrodef{PNS}[PNS]{Peripheral Nervous System}
\acrodef{PPG}[PPG]{Photoplethysmography}
\acrodef{PR}[PR]{Production Rule}
\acrodef{PSC}[PSC]{Post-Synaptic Current}
\acrodef{PSD}[PSD]{Power spectral density}
\acrodef{PSP}[PSP]{Post-Synaptic Potential}
\acrodef{PSTH}[PSTH]{Peri-Stimulus Time Histogram}
\acrodef{PV}[PV]{Parvalbumin}
\acrodef{PWM}[PWM]{Pulse-Width Modulation} 
\acrodef{QDI}[QDI]{Quasi Delay Insensitive}
\acrodef{RAM}[RAM]{Random Access Memory}
\acrodef{RA}[RA]{Resected Area}
\acrodef{RDF}[RDF]{Random Dopant Fluctuation}
\acrodef{RELU}[ReLu]{Rectified Linear Unit}
\acrodef{RLS}[RLS]{Recursive Least-Squares}
\acrodef{RMSE}[RMSE]{Root Mean Square-Error}
\acrodef{RRMSE}[RRMSE]{Relative Root Mean Square Error}
\acrodef{RMS}[RMS]{Root Mean Square}
\acrodef{RNN}[RNN]{Recurrent Neural Network}
\acrodef{ROLLS}[ROLLS]{Reconfigurable On-Line Learning Spiking}
\acrodef{RRAM}[R-RAM]{Resistive Random Access Memory}
\acrodef{R}[R]{Ripple}
\acrodef{RBF}[RBF]{Radial basis function}
\acrodef{RISC}[RISC]{Reduced Instruction Set Computer}
\acrodef{RSA}[RSA]{Respiratory Sinus Arrhythmia}
\acrodef{SAC}[SAC]{Selective Attention Chip}
\acrodef{SAT}[SAT]{Boolean Satisfiability Problem}
\acrodef{SCI}[SCI]{Spinal Cord Injury}
\acrodef{SCX}[SCX]{Silicon CorteX}
\acrodef{SD}[SD]{Signal-Driven}
\acrodef{SEM}[SEM]{Spike-based Expectation Maximization}
\acrodef{SCR}[SCR]{Skin Conductance Response}
\acrodef{SLAM}[SLAM]{Simultaneous Localization and Mapping}
\acrodef{SNN}[SNN]{Spiking Neural Network}
\acrodef{SNR}[SNR]{Signal to Noise Ratio}
\acrodef{SOC}[SoC]{System-On-Chip}
\acrodef{SOI}[SOI]{Silicon on Insulator}
\acrodef{SOZ}[SOZ]{Seizure Onset Zone}
\acrodef{SP}[SP]{Separation Property}
\acrodef{SPI}[SPI]{Serial Peripheral Interface}
\acrodef{SRAM}[SRAM]{Static Random Access Memory}
\acrodef{SST}[SST]{Somatostatin}
\acrodef{STDP}[STDP]{Spike-Timing Dependent Plasticity}
\acrodef{STD}[STD]{Short-Term Depression}
\acrodef{STP}[STP]{Short-Term Plasticity}
\acrodef{STT-MRAM}[STT-MRAM]{Spin-Transfer Torque Magnetic Random Access Memory}
\acrodef{STT}[STT]{Spin-Transfer Torque}
\acrodef{SVM}[SVM]{Support Vector Machine}
\acrodef{SW}[SW]{Software}
\acrodef{sWTA}[sWTA]{soft Winner-Take-All}
\acrodef{TEMP}[TEMP]{Temperature}
\acrodef{TCAM}[TCAM]{Ternary Content-Addressable Memory}
\acrodef{TFT}[TFT]{Thin Film Transistor}
\acrodef{TIME}[TIME]{Transverse Intrafascicular Multichannel Electrode}
\acrodef{TLE}[TLE]{Temporal Lobe Epilepsy}
\acrodef{UEA}[UEA]{Utah Electrode Array}
\acrodef{USB}[USB]{Universal Serial Bus}
\acrodef{USEA}[USEA]{Utah Slanted Electrode Array}
\acrodef{VHDL}[VHDL]{VHSIC Hardware Description Language}
\acrodef{VHSIC}[VHSIC]{Very High Speed Integrated Circuits}
\acrodef{VIP}[VIP]{Vasoactive Intestinal Peptide}
\acrodef{VLSI}[VLSI]{Very Large Scale Integration}
\acrodef{VNS}[VNS]{Vagal Nerve Stimulation}
\acrodef{VOR}[VOR]{Vestibulo-Ocular Reflex}
\acrodef{VSA}[VSA]{Vector Symbolic Architecture}
\acrodef{WCST}[WCST]{Wisconsin Card Sorting Test}
\acrodef{WSTD}[WSTD]{windowed standard deviation} 
\acrodef{WTA}[WTA]{Winner-Take-All}
\acrodef{XML}[XML]{eXtensible Mark-up Language}
\acrodef{ADC}[ADC]{Analog-to-Digital Converter}
\acrodef{ADEXP}[AdExp-IF]{Adaptive Exponential Integrate-and-Fire}
\acrodef{ADM}[ADM]{Asynchronous Delta Modulator}
\acrodef{AE}[AE]{Address-Event}
\acrodef{AER}[AER]{Address-Event Representation}
\acrodef{AEX}[AEX]{AER EXtension board}
\acrodef{AFE}[AFE]{Analog Front-End}
\acrodef{AFM}[AFM]{Atomic Force Microscope}
\acrodef{AGC}[AGC]{Automatic Gain Control}
\acrodef{AI}[AI]{Artificial Intelligence}
\acrodef{alphaMN}[\( \alpha \)-MN]{$\alpha$-motor neuron}
\acrodef{AMDA}[AMDA]{AER Motherboard with D/A converters}
\acrodef{AMPA}[AMPA]{$\alpha$-Amino-3-hydroxy-5-methyl-4-isoxazolepropionic Acid}
\acrodef{ANN}[ANN]{Artificial Neural Network}
\acrodef{API}[API]{Application Programming Interface}
\acrodef{APMOM}[APMOM]{Alternate Polarity Metal On Metal}
\acrodef{ARM}[ARM]{Advanced RISC Machine}
\acrodef{ASIC}[ASIC]{Application Specific Integrated Circuit}
\acrodef{BCM}[BMC]{Bienenstock-Cooper-Munro}
\acrodef{BD}[BD]{Bundled Data}
\acrodef{BEOL}[BEOL]{Back-end of Line}
\acrodef{BG}[BG]{Bias Generator}
\acrodef{BMI}[BMI]{Brain-Machince Interface}
\acrodef{BTB}[BTB]{Band-to-Band tunnelling}
\acrodef{bpm}[bpm]{Beats per Minute}
\acrodef{CA}[CA]{Cortical Automaton}
\acrodef{CAD}[CAD]{Computer Aided Design}
\acrodef{CAM}[CAM]{Content Addressable Memory}
\acrodef{CAVIAR}[CAVIAR]{Convolution AER Vision Architecture for Real-Time}
\acrodef{CCN}[CCN]{Cooperative and Competitive Network}
\acrodef{CDR}[CDR]{Clock-Data Recovery}
\acrodef{CFC}[CFC]{Current to Frequency Converter}
\acrodef{CHP}[CHP]{Communicating Hardware Processes}
\acrodef{CMIM}[CMIM]{Metal-Insulator-Metal Capacitor}
\acrodef{CML}[CML]{Current Mode Logic}
\acrodef{CMOL}[CMOL]{Hybrid CMOS nanoelectronic circuits}
\acrodef{CMOS}[CMOS]{Complementary Metal-Oxide-Semiconductor}
\acrodef{CNN}[CNN]{Convolutional Neural Network}
\acrodef{CNS}[CNS]{central Nervous System}
\acrodef{COTS}[COTS]{Commercial Off-The-Shelf}
\acrodef{CPG}[CPG]{Central Pattern Generator}
\acrodef{CPLD}[CPLD]{Complex Programmable Logic Device}
\acrodef{CPU}[CPU]{Central Processing Unit}
\acrodef{CSM}[CSM]{Cortical State Machine}
\acrodef{CSP}[CSP]{Constraint Satisfaction Problem}
\acrodef{CTXCTL}[CTXCTL]{CortexControl}
\acrodef{CV}[CV]{Coefficient of Variation}
\acrodef{DAC}[DAC]{Digital to Analog Converter}
\acrodef{DAS}[DAS]{Dynamic Auditory Sensor}
\acrodef{DAVIS}[DAVIS]{Dynamic and Active Pixel Vision Sensor}
\acrodef{DBN}[DBN]{Deep Belief Network}
\acrodef{DBS}[DBS]{Deep Brain Stimulation}
\acrodef{DFA}[DFA]{Deterministic Finite Automaton}
\acrodef{DIBL}[DIBL]{Drain-Induced Barrier-Lowering}
\acrodef{DI}[DI]{Delay Insensitive}
\acrodef{divmod3}[DIVMOD3]{Divisibility of a number by three}
\acrodef{DMA}[DMA]{Direct Memory Access}
\acrodef{DNF}[DNF]{Dynamic Neural Field}
\acrodef{DNN}[DNN]{Deep Neural Network}
\acrodef{DoA}[DoA]{Degree of Actuation}
\acrodef{DoF}[DoF]{Degree of Freedom}
\acrodef{DPE}[DPE]{Dynamic Parameter Estimation}
\acrodef{DPI}[DPI]{Differential Pair Integrator}
\acrodef{DRAM}[DRAM]{Dynamic Random Access Memory}
\acrodef{DR}[DR]{Dual Rail}
\acrodef{DRRZ}[DR-RZ]{Dual-Rail Return-to-Zero}
\acrodef{DSP}[DSP]{Digital Signal Processor}
\acrodef{DVS}[DVS]{Dynamic Vision Sensor}
\acrodef{DYNAP-SE}[DYNAP-SE]{Dynamic Neuromorphic Asynchronous Processor}
\acrodef{EBL}[EBL]{Electron Beam Lithography}
\acrodef{ECG}[ECG]{Electrocardiography}
\acrodef{ECoG}[ECoG]{Electrocorticography}
\acrodef{EDA}[EDA]{Electrodermal activity}
\acrodef{EDVAC}[EDVAC]{Electronic Discrete Variable Automatic Computer}
\acrodef{EEG}[EEG]{Electroencephalography}
\acrodef{EI}[EI]{Excitatory-Inhibitory}
\acrodef{EIN}[EIN]{Excitatory-Inhibitory Network}
\acrodef{EM}[EM]{Expectation Maximization}
\acrodef{EMG}[EMG]{Electromyography}
\acrodef{EOG}[EOG]{Electrooculogram}
\acrodef{EPSC}[EPSC]{Excitatory Post-Synaptic Current}
\acrodef{EPSP}[EPSP]{Excitatory Post-Synaptic Potential}
\acrodef{EZ}[EZ]{Epileptogenic Zone}
\acrodef{FDSOI}[FDSOI]{Fully-Depleted Silicon on Insulator}
\acrodef{FET}[FET]{Field-Effect Transistor}
\acrodef{FFT}[FFT]{Fast Fourier Transform}
\acrodef{FI}[F-I]{Frequency--Current}
\acrodef{FMA}[FMA]{Floating Microelectrode Array}
\acrodef{FNN}[FNN]{Feed-forward Neural Network}
\acrodef{FPGA}[FPGA]{Field Programmable Gate Array}
\acrodef{FR}[FR]{Fast Ripple}
\acrodef{FSA}[FSA]{Finite State Automaton}
\acrodef{FSM}[FSM]{Finite State Machine}
\acrodef{GABA}[GABA]{$\gamma$-Aminobutanoic Acid}
\acrodef{gammaMN}[\( \gamma \)-MN]{$\gamma$-motor neuron}
\acrodef{GIDL}[GIDL]{Gate-Induced Drain Leakage}
\acrodef{GOPS}[GOPS]{Giga-Operations per Second}
\acrodef{GPIO}[GPIO]{General Purpose I/O}
\acrodef{GPU}[GPU]{Graphical Processing Unit}
\acrodef{GT}[GT]{Ground Truth}
\acrodef{GUI}[GUI]{Graphical User Interface}
\acrodef{HAL}[HAL]{Hardware Abstraction Layer}
\acrodef{HD-sEMG}[HD-sEMG]{high-density surface electromyography}
\acrodef{HFO}[HFO]{High Frequency Oscillation}
\acrodef{HH}[H\&H]{Hodgkin \& Huxley}
\acrodef{HMI}[HMI]{Human Machine Interface}
\acrodef{HMM}[HMM]{Hidden Markov Model}
\acrodef{HR}[HR]{Heart Rate}
\acrodef{HRS}[HRS]{High-Resistive State}
\acrodef{HSD}[HSD]{Honest Significant Difference} 
\acrodef{HSE}[HSE]{Handshaking Expansion}
\acrodef{HW}[HW]{Hardware}
\acrodef{hWTA}[hWTA]{Hard Winner-Take-All}
\acrodef{HRV}[HRV]{hearth rate variability}
\acrodef{IC}[IC]{Integrated Circuit}
\acrodef{ICT}[ICT]{Information and Communication Technology}
\acrodef{iEEG}[iEEG]{Intracranial Electroencephalography}
\acrodef{IF2DWTA}[IF2DWTA]{Integrate \& Fire 2-Dimensional WTA}
\acrodef{IF}[I\&F]{Integrate-and-Fire}
\acrodef{IFSLWTA}[IFSLWTA]{Integrate \& Fire Stop Learning WTA}
\acrodef{IMU}[IMU]{Inertial Measurement Unit}
\acrodef{INCF}[INCF]{International Neuroinformatics Coordinating Facility}
\acrodef{INI}[INI]{Institute of Neuroinformatics}
\acrodef{IO}[I/O]{Input/Output}
\acrodef{IoT}[IoT]{Internet of Things}
\acrodef{IP}[IP]{Intellectual Property}
\acrodef{IPSC}[IPSC]{Inhibitory Post-Synaptic Current}
\acrodef{IPSP}[IPSP]{Inhibitory Post-Synaptic Potential}
\acrodef{ISI}[ISI]{Inter-Spike Interval}
\acrodef{JFLAP}[JFLAP]{Java - Formal Languages and Automata Package}
\acrodef{LEDR}[LEDR]{Level-Encoded Dual-Rail}
\acrodef{LFP}[LFP]{Local Field Potential}
\acrodef{LIFE}[LIFE]{Longitudinal Intrafascicular Electrodes}
\acrodef{LIF}[LIF]{Leaky Integrate-and-Fire}
\acrodef{LLC}[LLC]{Low Leakage Cell}
\acrodef{LMS}[LMS]{Least Mean Squares}
\acrodef{LNA}[LNA]{Low-Noise Amplifier}
\acrodef{LPF}[LPF]{Low Pass Filter}
\acrodef{LR}[LR]{Logistic Regression}
\acrodef{LRS}[LRS]{Low-Resistive State}
\acrodef{LSM}[LSM]{Liquid State Machine}
\acrodef{LTD}[LTD]{Long Term Depression}
\acrodef{LTI}[LTI]{Linear Time-Invariant}
\acrodef{LTP}[LTP]{Long Term Potentiation}
\acrodef{LTU}[LTU]{Linear Threshold Unit}
\acrodef{LUT}[LUT]{Look-Up Table}
\acrodef{LVDS}[LVDS]{Low Voltage Differential Signaling}
\acrodef{MCMC}[MCMC]{Markov-Chain Monte Carlo}
\acrodef{MAE}[MAE]{Mean Absolute Error}
\acrodef{MEA}[MEA]{Multielectrode Arrays}
\acrodef{MEMS}[MEMS]{Micro Electro Mechanical System}
\acrodef{MFR}[MFR]{Mean Firing Rate}
\acrodef{MIM}[MIM]{Metal Insulator Metal}
\acrodef{ML}[ML]{Machine Learning}
\acrodef{MLP}[MLP]{Multilayer Perceptron}
\acrodef{MSE}[MSE]{mean squared error}
\acrodef{monoNSM}[monoNSM]{Monotonic Neural State Machine}
\acrodef{MOSCAP}[MOSCAP]{Metal Oxide Semiconductor Capacitor}
\acrodef{MOSFET}[MOSFET]{Metal Oxide Semiconductor Field-Effect Transistor}
\acrodef{MOS}[MOS]{Metal Oxide Semiconductor}
\acrodef{MRI}[MRI]{Magnetic Resonance Imaging}
\acrodef{MU}[MU]{motor unit} 
\acrodef{MVC}[MVC]{Maximum Voluntary Contraction}
\acrodef{NCS}[NCS]{Neuromorphic Cognitive Systems}
\acrodef{NDFSM}[NDFSM]{Non-deterministic Finite State Machine}
\acrodef{ND}[ND]{Noise-Driven}
\acrodef{NEF}[NEF]{Neural Engineering Framework}
\acrodef{NHML}[NHML]{Neuromorphic Hardware Mark-up Language}
\acrodef{NIL}[NIL]{Nano-Imprint Lithography}
\acrodef{NI}[NI]{Neural Interface}
\acrodef{NMDA}[NMDA]{\textit{N}-Methyl-\textsc{d}-aspartate}
\acrodef{NME}[NE]{Neuromorphic Engineering}
\acrodef{NN}[NN]{Neural Network}
\acrodef{nnNSM}[nnNSM]{Nearest Neighbors Neural State Machine}
\acrodef{NOC}[NoC]{Network-on-Chip}
\acrodef{NRZ}[NRZ]{Non-Return-to-Zero}
\acrodef{NSM}[NSM]{Neural State Machine}
\acrodef{OLS}[OLS]{Ordinary Least Squares}
\acrodef{OR}[OR]{Operating Room}
\acrodef{OTA}[OTA]{Operational Transconductance Amplifier}
\acrodef{PCB}[PCB]{Printed Circuit Board}
\acrodef{PCHB}[PCHB]{Pre-Charge Half-Buffer}
\acrodef{PCM}[PCM]{Phase Change Memory}
\acrodef{PC}[PC]{Personal Computer}
\acrodef{PDK}[PDK]{Process Design Kit}
\acrodef{PE}[PE]{Phase Encoding}
\acrodef{PFA}[PFA]{Probabilistic Finite Automaton}
\acrodef{PFC}[PFC]{Prefrontal Cortex}
\acrodef{PFM}[PFM]{Pulse Frequency Modulation}
\acrodef{PGA}[PGA]{Programmable Gain Amplifier}
\acrodef{PID}[PID]{Proportional–integral–derivative}
\acrodef{PNI}[PNI]{Peripheral Nerve Interface}
\acrodef{PNS}[PNS]{Peripheral Nervous System}
\acrodef{PPG}[PPG]{Photoplethysmography}
\acrodef{PR}[PR]{Production Rule}
\acrodef{PSC}[PSC]{Post-Synaptic Current}
\acrodef{PSD}[PSD]{Power spectral density}
\acrodef{PSP}[PSP]{Post-Synaptic Potential}
\acrodef{PSTH}[PSTH]{Peri-Stimulus Time Histogram}
\acrodef{PV}[PV]{Parvalbumin}
\acrodef{PWM}[PWM]{Pulse-Width Modulation} 
\acrodef{QDI}[QDI]{Quasi Delay Insensitive}
\acrodef{RAM}[RAM]{Random Access Memory}
\acrodef{RA}[RA]{Resected Area}
\acrodef{RDF}[RDF]{Random Dopant Fluctuation}
\acrodef{RELU}[ReLu]{Rectified Linear Unit}
\acrodef{RLS}[RLS]{Recursive Least-Squares}
\acrodef{RMSE}[RMSE]{Root Mean Square-Error}
\acrodef{RRMSE}[RRMSE]{Relative Root Mean Square Error}
\acrodef{RMS}[RMS]{Root Mean Square}
\acrodef{RNN}[RNN]{Recurrent Neural Network}
\acrodef{ROLLS}[ROLLS]{Reconfigurable On-Line Learning Spiking}
\acrodef{RRAM}[R-RAM]{Resistive Random Access Memory}
\acrodef{R}[R]{Ripple}
\acrodef{RBF}[RBF]{Radial basis function}
\acrodef{RISC}[RISC]{Reduced Instruction Set Computer}
\acrodef{RSA}[RSA]{Respiratory Sinus Arrhythmia}
\acrodef{SAC}[SAC]{Selective Attention Chip}
\acrodef{SAT}[SAT]{Boolean Satisfiability Problem}
\acrodef{SCI}[SCI]{Spinal Cord Injury}
\acrodef{SCX}[SCX]{Silicon CorteX}
\acrodef{SD}[SD]{Signal-Driven}
\acrodef{SEM}[SEM]{Spike-based Expectation Maximization}
\acrodef{SCR}[SCR]{Skin Conductance Response}
\acrodef{SLAM}[SLAM]{Simultaneous Localization and Mapping}
\acrodef{SNN}[SNN]{Spiking Neural Network}
\acrodef{SNR}[SNR]{Signal to Noise Ratio}
\acrodef{SOC}[SoC]{System-On-Chip}
\acrodef{SOI}[SOI]{Silicon on Insulator}
\acrodef{SOZ}[SOZ]{Seizure Onset Zone}
\acrodef{SP}[SP]{Separation Property}
\acrodef{SPI}[SPI]{Serial Peripheral Interface}
\acrodef{SRAM}[SRAM]{Static Random Access Memory}
\acrodef{SST}[SST]{Somatostatin}
\acrodef{STDP}[STDP]{Spike-Timing Dependent Plasticity}
\acrodef{STD}[STD]{Short-Term Depression}
\acrodef{STP}[STP]{Short-Term Plasticity}
\acrodef{STT-MRAM}[STT-MRAM]{Spin-Transfer Torque Magnetic Random Access Memory}
\acrodef{STT}[STT]{Spin-Transfer Torque}
\acrodef{SVM}[SVM]{Support Vector Machine}
\acrodef{SW}[SW]{Software}
\acrodef{sWTA}[sWTA]{soft Winner-Take-All}
\acrodef{TEMP}[TEMP]{Temperature}
\acrodef{TCAM}[TCAM]{Ternary Content-Addressable Memory}
\acrodef{TFT}[TFT]{Thin Film Transistor}
\acrodef{TIME}[TIME]{Transverse Intrafascicular Multichannel Electrode}
\acrodef{TLE}[TLE]{Temporal Lobe Epilepsy}
\acrodef{UEA}[UEA]{Utah Electrode Array}
\acrodef{USB}[USB]{Universal Serial Bus}
\acrodef{USEA}[USEA]{Utah Slanted Electrode Array}
\acrodef{VHDL}[VHDL]{VHSIC Hardware Description Language}
\acrodef{VHSIC}[VHSIC]{Very High Speed Integrated Circuits}
\acrodef{VIP}[VIP]{Vasoactive Intestinal Peptide}
\acrodef{VLSI}[VLSI]{Very Large Scale Integration}
\acrodef{VNS}[VNS]{Vagal Nerve Stimulation}
\acrodef{VOR}[VOR]{Vestibulo-Ocular Reflex}
\acrodef{VSA}[VSA]{Vector Symbolic Architecture}
\acrodef{WCST}[WCST]{Wisconsin Card Sorting Test}
\acrodef{WSTD}[WSTD]{windowed standard deviation} 
\acrodef{WTA}[WTA]{Winner-Take-All}
\acrodef{XML}[XML]{eXtensible Mark-up Language}

\begin{document}

\title{Finger Force Decoding from Motor Units Activity on Neuromorphic Hardware}

\author{\IEEEauthorblockN{Farah Baracat\IEEEauthorrefmark{2}\IEEEauthorrefmark{1},
Giacomo Indiveri\IEEEauthorrefmark{2}, and 
Elisa Donati\IEEEauthorrefmark{2}%
}
\vspace{0.2cm}
\IEEEauthorblockA{\IEEEauthorrefmark{1}Corresponding author}
\IEEEauthorblockA{\IEEEauthorrefmark{2}University of Z{\"u}rich, and  ETH Z{\"u}rich, Z{\"u}rich, Switzerland--- \{fbarac, giacomo, elisa\}@ini.uzh.ch}
}

\maketitle

\begin{abstract}
Accurate finger force estimation is critical for next-generation human-machine interfaces. Traditional electromyography (EMG)-based decoding methods using deep learning require large datasets and high computational resources, limiting their use in real-time, embedded systems. Here, we propose a novel approach that performs finger force regression using spike trains from individual motor neurons, extracted from high-density EMG. These biologically grounded signals drive a spiking neural network implemented on a mixed-signal neuromorphic processor. Unlike prior work that encodes EMG into events, our method exploits spike timing on motor units to perform low-power, real-time inference. This is the first demonstration of motor neuron-based continuous regression computed directly on neuromorphic hardware. Our results confirm accurate finger-specific force prediction with minimal energy use, opening new possibilities for embedded decoding in prosthetics and wearable neurotechnology.
\end{abstract}


\section{Introduction}
\label{sec:intro}
Surface \ac{EMG} captures electrical signals generated by contracting muscles, providing a non-invasive window into the motor intent and enabling applications ranging from prosthetic control to gesture-based interaction in smart environments and virtual reality~\cite{parker_myoelectric_1986,lukyanenko_stable_2021,lin_vr-based_2023}. These signals have been used to estimate individual finger trajectories~\cite{chen_semg-based_2021}. However, most existing approaches rely on complex decoder architectures that demand large amounts of training data—often unavailable in real-world scenarios—and high computational resources, which can compromise system responsiveness and energy efficiency.

An alternative strategy for decoding motor intent involves extracting motor neuron activity directly from \ac{EMG}, yielding a sparse and physiologically meaningful representation of the neural drive~\cite{holobar_multichannel_2007,Grison_etal24}. By shifting from continuous muscle signals to discrete \ac{MU} discharges, this approach improves regression accuracy and simplifies the design of downstream decoders. Notably, the spiking nature of \ac{MU} activity aligns naturally with neuromorphic systems, which emulate the structure and dynamics of biological neural networks in silicon. These systems, built using mixed-signal analog/digital circuits in standard \ac{CMOS} technology, rely on \acp{SNN} to process sparse, time-dependent events, offering low-latency and energy-efficient computation ideally suited for real-time applications in resource-constrained settings~\cite{Mead90, Chicca_etal14}. 

Neuromorphic hardware has already demonstrated its potential across various biomedical domains, including Electrocardiography (ECG) anomaly detection~\cite{Bauer_etal19,DeLuca_etal25}, \ac{EMG}-based classification~\cite{Donati_etal19,Ceolini_etal20}, and~\ac{EEG}-driven biomarker extraction~\cite{Sharifshazileh_etal21, Gallou_etal24}. Here, we extend its application to the continuous estimation of finger forces. Previous studies have proposed event-based decoding schemes compatible with neuromorphic processing, including spike-based inference on embedded microcontrollers~\cite{zanghieri_regression, zanghieri2024event}, and encoding strategies to convert analog muscle signals into spikes for downstream \ac{SNN} use~\cite{Baracat_etal25}. However, these works have generally not demonstrated adaptable learning directly on mixed-signal neuromorphic chips in a closed-loop configuration.

In this work, we introduce a spiking-based framework for real-time finger force regression from \ac{MU} activity on the \ac{DYNAP-SE}, a mixed-signal neuromorphic processor. Spike trains corresponding to individual \acp{MU} are extracted from \ac{HD-sEMG} \ac{Hyser} dataset~\cite{Jiang_etal21}, and then streamed directly to the chip for decoding. Since the \ac{DYNAP-SE} does not support on-chip learning, we employ a computer-in-the-loop training strategy in which synaptic weights are iteratively updated based on the recorded output activity~\cite{Zhao_etal23}. This training paradigm allows the network to adapt to neuron variability and analog mismatch, while preserving the low-latency and low-power benefits of on-chip inference.

Our neuromorphic decoding framework adopts a two-stage architecture: (1) \ac{MU} decomposition from \ac{HD-sEMG}, and (2) real-time, spike-based decoding using a compact and scalable \ac{SNN}. This design yields an interpretable, low-power solution suitable for wearable applications, enabling embedded decoding in neuroprosthetics, assistive robotics, and immersive virtual environments. Once trained, the system performs finger-specific force estimation entirely on-chip, achieving a \ac{RMSE} of $7.85 \pm 1.18$,\% \ac{MVC} across all five fingers—demonstrating the feasibility of energy-efficient neuromorphic decoding for real-time human–machine interfacing.
\begin{figure*}[h!]
    \begin{subfigure}[b]{0.27\linewidth}
    {\includegraphics[width=1\linewidth]{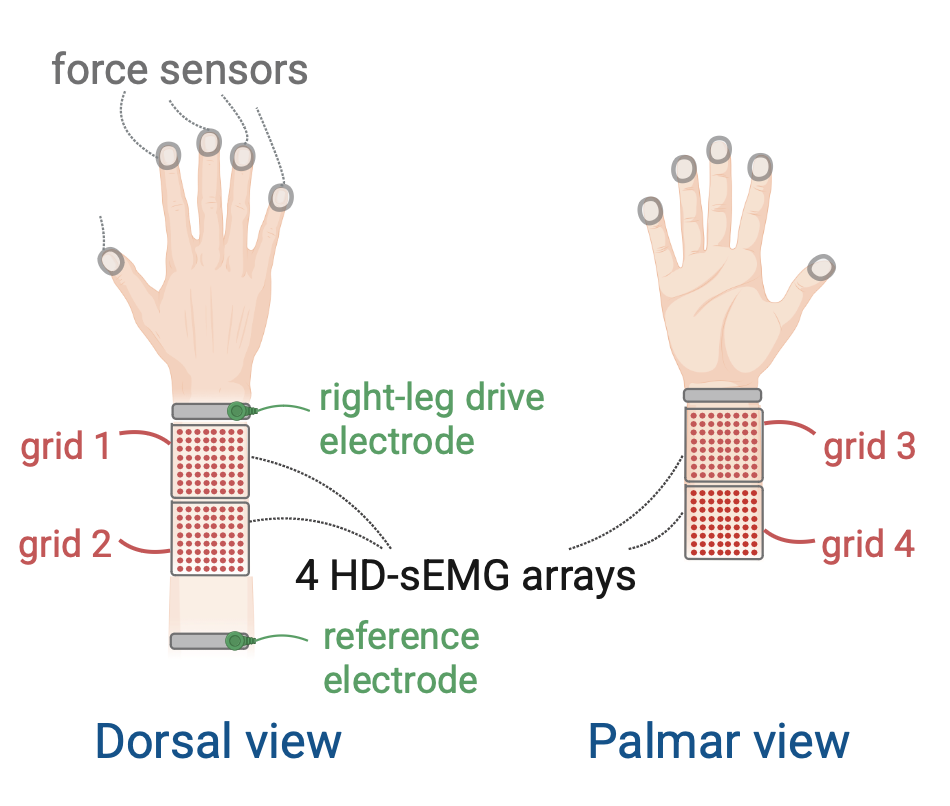}}
    \caption{}
    \label{fig:hyser_setup}
    \end{subfigure}
    \hfill
    \begin{subfigure}[b]{0.7\linewidth}
    {\includegraphics[width=1\linewidth]{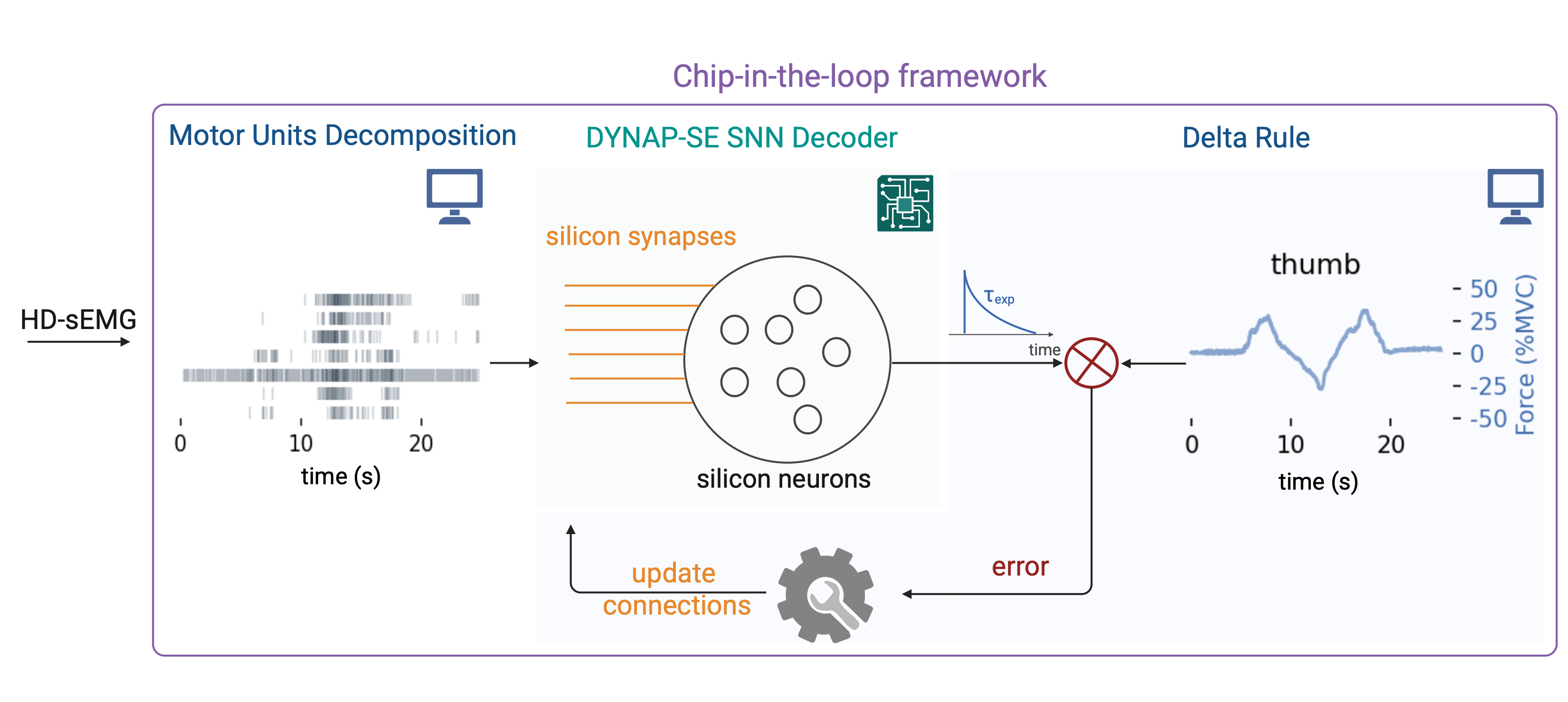}}
    \caption{}
    \label{fig:chip_in_loop}
    \end{subfigure}
    \caption{Overview of finger force decoder on \ac{DYNAP-SE}. (a) \ac{Hyser} experimental setup. Four 64-channel \ac{HD-sEMG} grids are placed on the forearm muscles with simultaneous individual finger force recording. (b) \ac{SNN} on \ac{DYNAP-SE} with connectivity trained within a computer-in-the-loop framework.}
\end{figure*}
\section{Material \& Methods}
\label{sec:MM}
\subsection{HD-sEMG Dataset}
\label{ssec:dataset}
In this work, we used the publicly available \ac{Hyser} dataset, which comprises \ac{HD-sEMG} recordings from the forearm muscles of 20 healthy participants, collected using four 64-channel grids for a total of 256 \ac{EMG} channels~\cite{Jiang_etal21}. Simultaneously, ground-truth force trajectories of the five fingers were recorded using five pairs of force sensors, with each pair positioned to independently measure flexion and extension forces for a given digit (see Fig.~\ref{fig:hyser_setup}). We focused on the one-\ac{DoF} sub-dataset, which involves isolated isometric contractions of individual fingers following a triangular force profile with peak amplitudes of 30\% \ac{MVC} in both flexion and extension. For each finger, three trials were collected, with each trial lasting 25 seconds.

\subsection{Offline \ac{MU} Decomposition}
\label{ssec:mu_decomp}
\ac{HD-sEMG} signals were decomposed into individual \ac{MU} discharge patterns using the \ac{fastICA} algorithm with peel-off~\cite{negro_multi-channel_2016}, implemented in the open-source MUedit software~\cite{Avrillon_etal24}. The algorithm iteratively estimates a separation matrix that isolates the discharge timings of each \ac{MU} from the overlapping activity of other concurrently active units. Separation vectors were optimized on the first trial and subsequently fixed to track the same \acp{MU} across trials. For the remaining trials, \ac{EMG} signals were projected onto the learned separation matrix, and candidate peaks from the resulting \ac{MU} pulse trains were classified into spike and noise classes using K-means clustering~\cite{barsakcioglu_control_2021, rossato_i-spin_2024}. The reliability of each identified \ac{MU} was assessed using the silhouette score, with only \acp{MU} exceeding a threshold of 0.9 and exhibiting discharge rates within the physiological range of \SI{2}{Hz} to \SI{50}{Hz} retained for further analysis.

Decomposition was carried out separately for each finger task, resulting in finger-specific \ac{MU} sets. This approach aimed to maximize the number of identified \acp{MU} per finger and assess the feasibility of training a force decoder on neuromorphic hardware within a computer-in-the-loop framework. By providing an optimized input set, the setup served as a preliminary test case to validate decoder performance under high-quality \ac{MU} inputs, with variability constrained to inter-trial differences.

\subsection{Baseline Force Decoder}
\label{ssec:baseline_decoder}
For each finger task, a linear regressor was fitted to the decomposed \ac{MU} activity and served as a baseline model. \ac{MU} activity was segmented into \SI{100}{\milli\second} windows with 50\% overlap, and spike counts from each \ac{MU} were computed to form the input feature vectors. As with the decomposition, the ordinary least squares coefficients were estimated from the first trial and applied to subsequent trials for evaluation. Decoder performance was quantified using the \ac{RMSE} between the predicted and recorded force trajectories (in \%\ac{MVC}).

\subsection{Mixed-signal Neuromorphic Processor: DYNAP-SE}
\label{ssec:dynapse}
The neuromorphic implementation of \ac{MU} decoding was realized on the \ac{DYNAP-SE}, which emulates the biophysical properties of biological neurons and synapses~\cite{Moradi_etal18}. The chip comprises a total of 1024 analog neurons, distributed across four cores of 256 neurons each. The neuron's circuits implement the \ac{ADEXP} neuron model~\cite{Brette_Gerstner05}, and can be connected via four types of synapses--two excitatory (AMPA, NMDA) and two inhibitory (GABA-A, GABA-B)--with tunable temporal dynamics. Neurons and synapses within the same core share a common set of programmable parameters, which define their response properties; however, individual responses will differ due to transistor-level device mismatch, inherent in the analog circuits~\cite{Zendrikov_etal22}. External input spikes are handled by a dedicated \ac{FPGA} interface, which transmits address-events to the chip using the \ac{AER} protocol.

\subsection{Force Decoder on DYNAP-SE}
\label{ssec:chip_in_the_loop}
For each finger task, two populations of 20 neurons were allocated on the \ac{DYNAP-SE} on two separate cores-one designated for flexion and the other for extension. The populations received spike trains from decomposed \acp{MU}, with flexion-associated population driven by \acp{MU} identified from grids 1 and 2, and extension-associated population receiving input from grids 3 and 4 (see Fig.~\ref{fig:hyser_setup}).

The average firing rate across each neuronal population was used to estimate the corresponding force output. This population-level design mitigated neuron-to-neuron mismatch inherent in analog substrates, leveraging redundancy to improve decoding robustness. Decomposed \ac{MU} spike trains were streamed to the two populations on chip via excitatory AMPA synapses in an all-to-all feedforward configuration, forming a connectivity matrix between input \acp{MU} and output neurons on chip.

To train the connectivity, we employed a chip-in-the-loop learning strategy (illustrated in Fig.\ref{fig:chip_in_loop}), where updates to the synaptic matrix were computed off-chip based on the recorded output activity. Neuronal and synaptic circuit parameters were initialized at the beginning of training and remained fixed throughout. Rather than adjusting floating-point weights, the training procedure modified the number of active synapses between each input-output pair, thus encoding synaptic strength as discrete integers. Increasing the number of excitatory synapses enhanced the excitatory influence of a \ac{MU} on a given neuron, while reducing the synapse count—or substituting inhibitory synapses—attenuated its effect. In this way, training selectively rewired and reweighted the network to shape the population response toward accurate force decoding. The full procedure is outlined in Algorithm~\ref{alg:chip_loop}.

\begin{figure}[th]
    \centering
    {\includegraphics[width=.8\linewidth]{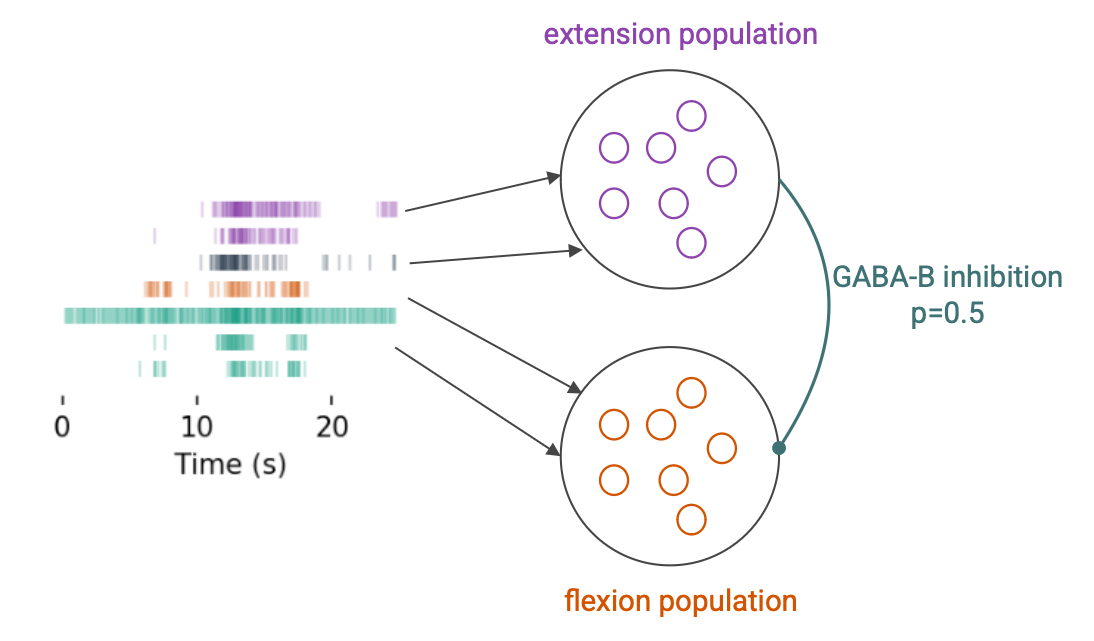}}
    \caption{\ac{DYNAP-SE} network topology during inference.}
     \label{fig:topology}
\end{figure}
During inference, we introduced additional untrained, unilateral inhibitory synapses from the extension to the flexion population (Fig.~\ref{fig:topology}) to partially compensate for the asymmetry in \ac{MU} counts—specifically, the greater number of flexion-associated \acp{MU} identified from grids 1 and 2. These inhibitory projections served to balance the network’s directional responsiveness but were excluded from the training procedure.

\begin{algorithm}
\caption{Computer-in-the-Loop Training Framework}
\label{alg:chip_loop}
\begin{algorithmic}[1]
\State \textbf{Given:} finger task $i$ with $N_{\text{MU}}$ \acp{MU}
\State \textbf{Allocate} $M_{\text{out}}$ output neurons on-chip
\State \textbf{Initialize} synaptic connectivity matrix $W \in \mathbb{Z}^{N_{\text{MU}} \times M_{\text{out}}}$
\State \textbf{Configure} neuron and synapse biases on \ac{DYNAP-SE}
\For{each training epoch}
    \State \textbf{Stream} \acp{MU} spike trains to the chip
    \State \textbf{Record} spiking activity of on-chip neurons
    \State \textbf{Estimate} instantaneous firing rates using an exponential kernel with decay, $\tau = \SI{200}{\milli\second}$
    \State \textbf{Compute} mean squared error between average population rate and ground truth force
    \State \textbf{Compute error gradient} and the updated weights $W' \in \mathbb{R}^{N_{\text{MU}} \times M_{\text{out}}}$
    \For{each $w'_{ij} \in W'$}
        \State Generate $r \sim \mathcal{U}(0,1)$
        \State Compute residual component $r_{ij} = w'_{ij} - \lfloor w'_{ij} \rfloor$
        \State \textbf{If} $r_{ij} > r$ \textbf{then} round $w'_{ij}$ up; \textbf{else} round $w'_{ij}$ down
        \State Assign $W_{ij} \in \{-k, \dots, -1, 0, +1, \dots, +k\}$
    \EndFor
    \State \textbf{Apply} updated connectivity matrix on-chip
\EndFor
\end{algorithmic}
\end{algorithm}

\begin{figure*}[th]
    {\includegraphics[width=1\linewidth]{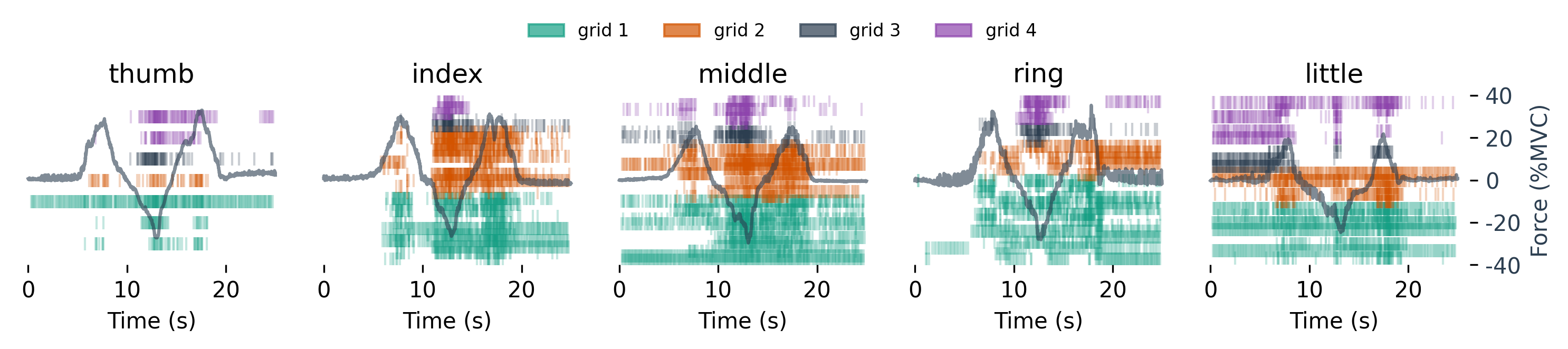}}
    \caption{Example decomposed \ac{MU} spike trains on trial 1 for the five fingers color-coded by the grid of origin. }
     \label{fig:decomposition}
\end{figure*}

\section{Results}
\label{sec:res}

\subsection{Motor Unit Decomposition}
\label{ssec:mu_consistency}
Figure~\ref{fig:decomposition} illustrates the decomposition of \ac{MU} activity during isolated single-finger contractions from Subject 1. A total of 132 \acp{MU} were identified across the five finger tasks. Spike trains were extracted from the four electrode arrays (color-coded: green for grid 1, orange for grid 2, black for grid 3, and purple for grid 4). The overlaid black trace represents the recorded force (\%\ac{MVC}) for each finger task. 

Distinct activation patterns emerge across fingers and recording grids. \acp{MU} extracted from grid 3 (black) and grid 4 (purple) exhibit selective activation during extension phases and are thus routed to the extension-related neuronal population on chip—most prominently for the thumb, index, and little fingers. In contrast, \acp{MU} from grid 1 (green) and grid 2 (orange) demonstrate broader activation across both flexion and extension, with a slight bias toward flexion. These units were input to the flexion-related population on chip.

\begin{figure*}[ht]
         {\includegraphics[width=1\linewidth]{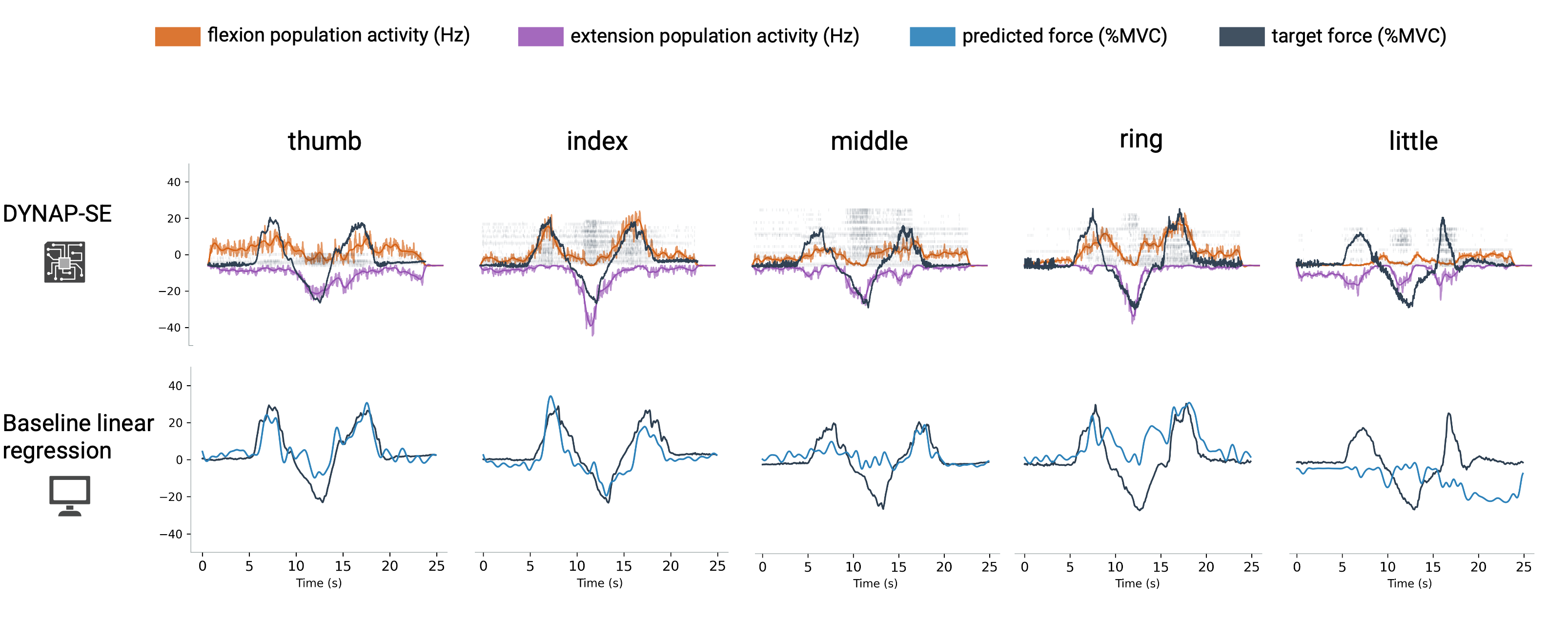}}
         \caption{Representative force decoding performance across all five fingers for Subject 1, trial 2. Top row: average population activity recorded from the \ac{DYNAP-SE}, using two complementary neuron populations—one responsive to flexion (orange) and the other to extension (purple). Bottom row: baseline linear regression using binned \ac{MU} spike counts as input features. The ground-truth trajectory (black) was also binned to match the temporal resolution of the predicted signal.}
         \label{fig:regression_predictions}
\end{figure*}

\subsection{Neuromorphic vs. Baseline Decoder}
\label{ssec:comparison}
Figure~\ref{fig:regression_predictions} presents representative force decoding results obtained from the \ac{DYNAP-SE} following training of the feedforward connectivity. For each finger, two separate neuron populations—one tuned to flexion and the other to extension—were used, and their mean firing rates served as the decoded force. To mitigate asymmetries in input \ac{MU} distributions, unidirectional inhibitory synapses were introduced from the extension to the flexion population during inference. These projections effectively reduced flexion population's activity during extension, as observed in its suppressed firing in the corresponding movement phase. Note that population activity, while strictly positive, is plotted with inverted sign for the extension phase to conform to the force signal convention.

Across testing trials 2 and 3, the neuromorphic decoder achieved average \ac{RMSE} values of $8.16 \pm 1.29~\%$ \ac{MVC} for the thumb, $8.62 \pm 2.07~\%$ for the index, $6.44 \pm 0.99~\%$ for the middle, $8.41 \pm 1.35~\%$ for the ring, and $7.66 \pm 0.20~\%$ for the little finger. These values represent the average of five runs to account for the stochasticity of the chip. In comparison, baseline linear regression models trained on binned \ac{MU} spike counts yielded \acp{RMSE} of $5.86 \pm 0.59~\%$, $5.42 \pm 1.15~\%$, $7.38 \pm 1.03~\%$, $12.16 \pm 1.29~\%$, and $20.20 \pm 5.72~\%$ \ac{MVC}  for the thumb, index, middle, ring, and little fingers, respectively. While the linear decoder exhibited slightly lower errors for the thumb and index fingers, its performance degraded considerably for the ring and little fingers. In contrast, the neuromorphic decoder achieved more consistent performance across all fingers and successfully captured both flexion and extension phases—for example, in the ring finger—highlighting the potential of on-chip decoding to match or even surpass conventional approaches at low power. Based on circuit-level estimates~\cite{Risi_etal20}, the power consumption during inference for our 40-neuron topology is $\SI{7.73}{\micro\watt}$, with mean output firing rates of $\SI{29.97}{\hertz}$ and $\SI{17.91}{\hertz}$ for the flexion and extension populations, respectively.

\section{Discussion and Conclusion}
\label{sec:conclusion}
We proposed an alternative approach for on-line finger-force decoding from \acp{MU} using an emerging technology that promises to enable low-power and low-latency sensory processing in resource-constrained setups. Rather than competing with conventional AI approaches that reach very good performances on available (off-line, digitized) datasets, we are proposing a complementary approach that exploits the inherent dynamics of its analog computing elements, resulting in neural networks that require far less parameters, memory and power of their machine learning counterparts. We validated this approach with an existing prototype chip, and demonstrated comparable performance to traditional methods from experiments with the real hardware. This study paves the way for a low-power, minimal-footprint neuromorphic solution for real-time EMG-based decoding of finger forces.

\printbibliography

\end{document}